%% file: main.tex
\documentclass{article}

\usepackage{natbib}
\usepackage[letterpaper,margin=1in]{geometry}

\usepackage[utf8]{inputenc}
\usepackage[T1]{fontenc}
\usepackage{amsmath}
\usepackage{amssymb}
\usepackage{graphicx}
\usepackage{booktabs}
\usepackage{array}
\usepackage{multirow}
\usepackage{microtype}
\usepackage{xcolor}
\usepackage{soul}
\sethlcolor{yellow}
\soulregister\cite7
\soulregister\citep7
\soulregister\ref7
\soulregister\ours7
\soulregister\textbf7
\soulregister\emph7

\usepackage{url}
\usepackage[hypertexnames=false]{hyperref}
\usepackage{algorithm}
\usepackage{algorithmic}
\usepackage{tikz}
\usetikzlibrary{arrows.meta,positioning,fit,calc,decorations.pathreplacing}
\usepackage{pifont}
\usepackage{makecell}
\usepackage[capitalise,noabbrev]{cleveref}
\newcommand{\cmark}{\ding{51}}
\newcommand{\xmark}{\ding{55}}

\newcommand{\ours}{RoboCoEvo}

\title{RoboCoEvo: Expanding Robot Capabilities through Agent--Policy Co-Evolution}

\author{
Ran Qi$^{1,*,\ddagger}$ \quad Yixin Xiao$^{2,*}$ \quad Dawei Wang$^{2}$ \quad Zhiyuan Li$^{1,\ddagger}$ \quad Jinbang Huang$^{2}$ \\
Xuan Zhao$^{2}$ \quad Yintao Ma$^{2}$ \quad Yangzheng Wu$^{2}$ \quad Tongtong Cao$^{3}$ \quad Yingxue Zhang$^{2,\dagger}$ \\[6pt]
$^{1}$University of Toronto \quad $^{2}$Huawei Noah's Ark Lab \\
$^{3}$Department of Foundation Model, 2012 Labs
}
\date{}

\begin{document}
\raggedbottom

\maketitle
\begin{NoHyper}
{\renewcommand{\thefootnote}{}
\footnotetext{$^{*}$Equal contribution.}
\footnotetext{$^{\dagger}$Correspondence to: \texttt{yingxue.zhang@huawei.com}.}
\footnotetext{$^{\ddagger}$Work done during an internship at Huawei Noah's Ark Lab.}}
\end{NoHyper}

\begin{abstract}
Integrating MLLM-based agents with robotic foundation models has attracted increasing attention as a means of enhancing robotic capabilities. However, the improvement of such integrated systems remains largely decoupled across their agent and model components, with advances in one component not necessarily benefiting the other, substantially limiting the capabilities of the combined system. 
As both MLLMs and robotic foundation models become increasingly capable, fully realizing their combined potential calls for co-evolution, an iterative process in which improvements in agents and robotic foundation models reinforce one another.
We propose \textbf{\ours{}}, a failure-guided co-evolution framework that couples the refinement of the agent's explicit and reusable executable knowledge with parameter learning in robotic foundation models, such as vision-language-action (VLA) models. Through a code-as-policy formulation, the agent develops and refines executable failure-detection and recovery programs,
generating targeted corrective experience.
Post-training consolidates the resulting corrective behaviors
into the robotic model's end-to-end policy, improving autonomous task
execution. Changes in the updated policy's failure distribution,
in turn, guide subsequent revisions to the agent's programs.
Experiments on BEHAVIOR-1K, RoboCasa, and RoboTwin
demonstrate sustained improvements in autonomous VLA task
performance across successive co-evolution rounds.
\end{abstract}

\input{sections/intro}

\input{sections/method}

\begingroup

\section{Experiments}
\label{sec:experiments}

We evaluate \ours{} on three well-known testbeds: BEHAVIOR-1K~\citep{li2024behavior1k}, RoboTwin~2.0~\citep{chen2026robotwin}, and RoboCasa~\citep{nasiriany2024robocasa}. In this section, we first describe our experimental setup, including benchmark selection, metrics, and training details, in Section~\ref{sec:setup}. We then present the main results of our method to demonstrate its efficiency and performance relative to the base policies. Finally, we conduct ablation studies to assess the effectiveness of each component of our method.

\subsection{Experimental Setup}
\label{sec:setup}

To comprehensively evaluate our method, we select three well-known simulation-based benchmarks:
\begin{itemize}
    \item BEHAVIOR-1K~\citep{li2024behavior1k} contains long-horizon, bimanual mobile manipulation tasks in household scenes and is one of the most challenging benchmarks for robotic manipulation. We use its first six tasks, covering the most common household activities.
    
    \item RoboTwin~2.0~\citep{chen2026robotwin} covers bimanual tabletop manipulation with diverse coordination requirements. We evaluate on all 50 tasks.
    
    \item RoboCasa~\citep{nasiriany2024robocasa} provides everyday manipulation tasks in realistic and diverse kitchen environments, using a single-arm mobile manipulator. We select ten tasks from different task categories, covering the benchmark's most representative task types. The selected tasks are listed in Appendix~\ref{app:robocasa-tasks}.
\end{itemize}

For each benchmark, we adopt a state-of-the-art policy as the starting point for self-improvement to demonstrate the effectiveness of our method in further advancing policy performance. For BEHAVIOR-1K, we use Robot Learning Collective (RLC)~\citep{larchenko2025behavior}, the first-place solution in the 2025 BEHAVIOR Challenge. For RoboTwin~2.0, we use $\pi_{0.5}$~\citep{black2025pi05} trained on both clean and randomized demonstrations, which serves as a strong VLA baseline on this benchmark with strong generalization capabilities. For RoboCasa, we use XR-1~\citep{fan2026xr1}, the top-ranked method as of September 2026. We use official checkpoints when available; otherwise, we use our reproduced models.

We report the task success rates on all benchmarks and, on BEHAVIOR-1K, also the challenge
Q-score~\citep{behavior2025challenge}, the fraction of goal predicates satisfied at the end of an episode. All policies are evaluated on the official test instances, with 50 instances per task on
RoboTwin and RoboCasa. On BEHAVIOR-1K, we run each of the 20 test instances per task twice. \ours{} improves a VLA that executes tasks on
its own, so we compare each updated policy with its base policy under the same evaluation protocol.

\paragraph{Data generation and training.}

In each round, we collect rollouts with the current policy using seeds distinct from those used for training and official testing. On BEHAVIOR-1K, we collect 50 rollouts per task and sample 20 failures. On RoboCasa and RoboTwin~2.0, we collect 200 rollouts per task to obtain a sufficient number of failures, as the base policies achieve higher success rates on these benchmarks than on BEHAVIOR-1K. For each FM, the agent writes a detector and a recovery program (Section~\ref{sec:discovery}). The execution framework described in Section~\ref{sec:method} then applies the detectors to all failed rollouts, runs the recovery programs on the detected failure states, and uses the agent to perturb these states for failure-conditioned augmentation (Section~\ref{sec:multiplication}). We use GPT-6 Astra with high reasoning effort as the agent. During fine-tuning of the VLA, we sample the original demonstrations and the accumulated failure experience at a 1:1 ratio to mitigate catastrophic forgetting. The training hyperparameters are provided in Appendix~\ref{app:training}.

\begin{table}[t]
\centering
\caption[]{\textbf{Policy performance and failure-mode analysis before and after \ours{}.} The first row of each benchmark is its starting policy. Bold marks the best Q-score and success rate per benchmark. The failure-mode columns are obtained by applying the failure-mode detectors to every test episode (Section~\ref{sec:results} and Appendix~\ref{app:fm-metrics}).}
\label{tab:main}
\small
\setlength{\tabcolsep}{3.5pt}
\begin{tabular}{@{}llccccccc@{}}
\toprule
& & & & \multicolumn{3}{c}{\textbf{Targeted failure modes}} & \multicolumn{2}{c}{\textbf{Failed episodes}} \\
\cmidrule(lr){5-7}\cmidrule(l){8-9}
\textbf{Benchmark} & \textbf{Policy} & Q-score & Succ. & FM occurs & Self-recovered & Fail $\mid$ FM & Targeted FM & Unexplained \\
\midrule
\multirow{3}{*}{BEHAVIOR-1K} & RLC     & 0.40 & 0.23 & 0.73 & 0.28 & 0.87 & 0.629 & 0.142 \\
                             & Round 1 & 0.49 & 0.29 & 0.69 & 0.44 & 0.78 & 0.542 & 0.167 \\
                             & Round 2 & \textbf{0.56} & \textbf{0.34} & 0.82 & 0.56 & 0.65 & 0.538 & 0.121 \\
\midrule
\multirow{3}{*}{RoboTwin}    & $\pi_{0.5}$ & -- & 0.79 & 0.29 & 0.18 & 0.65 & 0.188 & 0.021 \\
                             & Round 1 & -- & 0.90 & 0.16 & 0.54 & 0.38 & 0.060 & 0.040 \\
                             & Round 2 & -- & \textbf{0.93} & 0.19 & 0.54 & 0.31 & 0.058 & 0.017 \\
\midrule
\multirow{3}{*}{RoboCasa}    & XR-1    & -- & 0.79 & 0.30 & 0.35 & 0.59 & 0.176 & 0.028 \\
                             & Round 1 & -- & 0.88 & 0.22 & 0.46 & 0.45 & 0.099 & 0.014 \\
                             & Round 2 & -- & \textbf{0.91} & 0.24 & 0.62 & 0.31 & 0.077 & 0.014 \\
\bottomrule
\end{tabular}
\end{table}

\begin{table}[t]
\centering
\caption{\textbf{Per-task results on BEHAVIOR-1K.} Q-score and success rate of the base policy and
after each round of \ours{}. For each task and metric, the best result is in bold and the second best
underlined.}
\label{tab:b1k}
\small
\begin{tabular}{@{}lcccccc@{}}
\toprule
& \multicolumn{2}{c}{\textbf{Base}} & \multicolumn{2}{c}{\textbf{Round 1}} & \multicolumn{2}{c}{\textbf{Round 2}} \\
\cmidrule(lr){2-3}\cmidrule(lr){4-5}\cmidrule(l){6-7}
\textbf{Task} & Q & Succ. & Q & Succ. & Q & Succ. \\
\midrule
0 Turning on a radio & 0.45 & 0.45 & \underline{0.48} & \underline{0.48} & \textbf{0.70} & \textbf{0.70} \\
1 Picking up trash & 0.63 & 0.43 & \textbf{0.84} & \underline{0.58} & \underline{0.82} & \textbf{0.65} \\
2 Putting away Halloween decorations & 0.36 & 0.08 & \underline{0.47} & \underline{0.13} & \textbf{0.60} & \textbf{0.18} \\
3 Cleaning up plates and food & 0.40 & 0.23 & \underline{0.47} & \underline{0.33} & \textbf{0.52} & \textbf{0.35} \\
4 Canning meat & 0.02 & 0.00 & \underline{0.03} & 0.00 & \textbf{0.09} & 0.00 \\
5 Setting mousetraps & 0.54 & \underline{0.20} & \underline{0.63} & \textbf{0.25} & \textbf{0.66} & 0.18 \\
\bottomrule
\end{tabular}
\end{table}

\begin{figure}[t]
\centering
\includegraphics[width=0.98\linewidth]{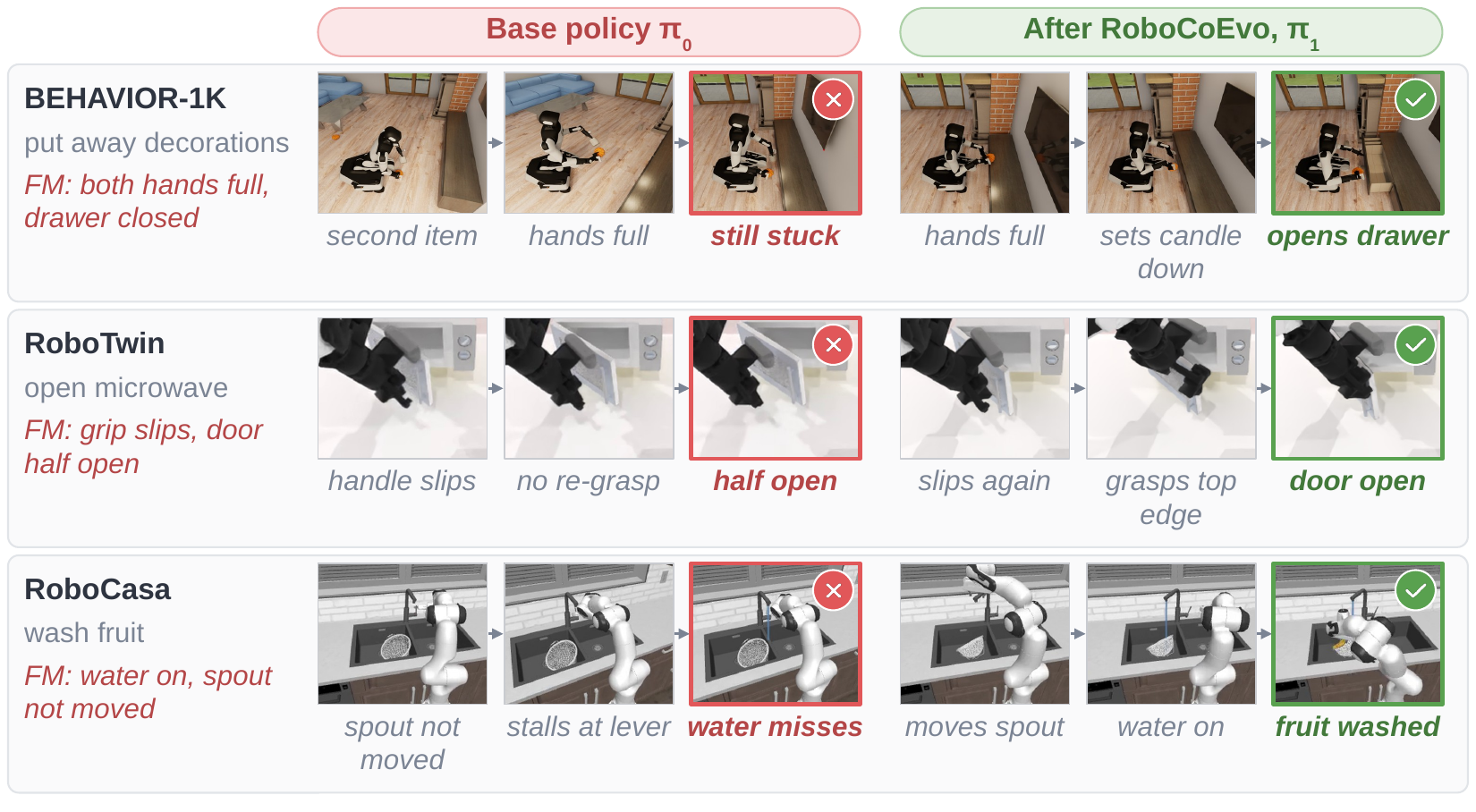}
\caption[]{\textbf{Qualitative comparison of the base policy and \ours{} on targeted failure modes.} Each row shows one failure mode on one benchmark. Left: the
base policy {$\pi_0$} enters the failure mode and fails to recover. Right: after \ours{}, {$\pi_1$} enters the same failure mode
and recovers without the agent. Frames are aligned by stage.}
\label{fig:qualitative}
\end{figure}

\subsection{Main Results}
\label{sec:results}

Overall, \ours{} improves the base policies across all three benchmarks, as shown in Table~\ref{tab:main}. %
 After two rounds, the average success rate rises from 79\% to 93\% on RoboTwin and from 79\% to 91\% on RoboCasa, and the mean Q-score on BEHAVIOR-1K increases from 0.40 to 0.56. These results show that failure-targeted experience can translate into improved full-task performance across different manipulation settings.

The gains develop over successive rounds of experience acquisition and policy updates. Both rounds improve the policy on every benchmark, and the first round yields the larger gain in success rate. These multi-round gains demonstrate \ours{}'s ability to improve through successive policy updates. As the policy becomes more capable, continued trial and error reveals remaining weaknesses and provides new targets for learning, suggesting a path toward a recursive improvement cycle of experience acquisition and capability growth. 

We also report per-task results on BEHAVIOR-1K to illustrate our method's performance across different tasks in Table~\ref{tab:b1k}. The results show steady Q-score improvements across both rounds on most tasks. On picking up trash, the Q-score rises from 0.63 to 0.84 after the first round, then decreases slightly to 0.82 in the second round. The strong first-round performance and limited subsequent gains suggest that learning on this task may be approaching saturation. Canning meat, by contrast, starts with a Q-score of 0.02 and gradually improves to 0.03 and then to 0.09. These small gains understate the progress: as the policy takes the jar out of the cabinet more reliably, more rollouts reach the later subtasks of opening the jar and fetching the sausage, where new failure modes are exposed (Appendix~\ref{app:fm-level}). Overall, these results support \ours{}'s ability to improve policies over multiple rounds, with the pace and extent of improvement varying according to the task and the policy's existing capabilities.

To understand these gains, we apply the failure-mode detectors to every test episode of each policy (Table~\ref{tab:main}). The base and first-round policies are evaluated using first-round detectors, and the second-round policy using second-round detectors. For each policy, we report the fraction of episodes in which a targeted failure mode occurs (\emph{FM occurs}), the fraction of occurrences from which the policy recovers and completes the step without assistance (\emph{Self-recovered}), and the fraction of episodes with an occurrence in which the failure mode is never recovered and the episode fails (\emph{Fail $\mid$ FM}). Failed episodes are further divided into those that fail with an unrecovered targeted failure mode (\emph{Targeted FM}) and those that fail without one (\emph{Unexplained}). Following the first round, the targeted failure modes occur less frequently, and the updated policy is more likely to recover autonomously when failures arise. On RoboTwin, the share of occurrences the policy recovers from rises from 0.18 to 0.54, and the fraction of episodes that fail by a targeted failure mode drops from 0.19 to 0.06. The policy has therefore learned to get out of these failures, not only to avoid them. Figure~\ref{fig:qualitative} shows one example per benchmark: the base policy stays in its failure, while the first-round policy enters the same kind of failure and recovers on its own. A per-failure-mode breakdown of these results is provided in Appendix~\ref{app:fm-level}.

The second-round experiment further highlights the importance of adapting detectors and recovery programs to each new policy (Table~\ref{tab:agent-evolution}).
When the first-round programs are applied without modification to the failed rollouts of the first-round policy, the detection success rate falls from at least 94\% to between 54\% and 72\%, and fewer than half of the failure states are detected and recovered. After the agent revises the programs using these rollouts, detection success returns to its first-round level, while recovery success increases to 71\%--83\%. The revised detectors also account for more of the failures of the second-round policy on the test episodes. On RoboTwin, for instance, the fraction of episodes that fail without an unrecovered targeted failure mode decreases from 0.040 for the first-round policy to 0.017 for the second-round policy. These results indicate that the policy's failure distribution shifts as it improves, such that programs developed for an earlier version no longer cover its current failures. Revising these programs in each round keeps the corrective experience targeted at the current policy's failures.

\begin{table}[t]
\centering
\caption{\textbf{Detectors and recovery programs on the first-round policy's failures.} Detection: fraction of failed rollouts attributed to the correct failure mode. Recovery: fraction of failure states restored from failed rollouts that are detected and then recovered by the recovery program. \emph{Round~1}: the first-round detectors and programs on the base policy's rollouts. \emph{Reused}: the same detectors and programs, unchanged, on the first-round policy's rollouts. \emph{Revised}: the detectors and programs after the agent's second-round revision, on the same rollouts.}
\label{tab:agent-evolution}
\small
\begin{tabular}{@{}lcccccc@{}}
\toprule
& \multicolumn{3}{c}{\textbf{Detection success}} & \multicolumn{3}{c}{\textbf{Recovery success}} \\
\cmidrule(lr){2-4}\cmidrule(l){5-7}
\textbf{Benchmark} & Round~1 & Reused & Revised & Round~1 & Reused & Revised \\
\midrule
BEHAVIOR-1K & 0.97 & 0.54 & 0.96 & 0.85 & 0.27 & 0.78 \\
RoboTwin    & 0.94 & 0.60 & 0.93 & 0.94 & 0.33 & 0.83 \\
RoboCasa    & 0.94 & 0.72 & 0.94 & 0.67 & 0.47 & 0.71 \\
\bottomrule
\end{tabular}
\end{table}

\begin{table}[t]
\centering
\caption{\textbf{Ablations on picking up trash (BEHAVIOR-1K).} Rec.: recovery data. FMs: shared failure-mode programs. Aug.: failure-conditioned augmentation. Tokens are given in millions (M), including cached input, over all agent sessions and the orchestrating session. GPU h: GPU hours of the base-policy rollouts used to obtain the failure states (for success-only data, the successful rollouts) and of recording the data, excluding development and testing. Performance is measured on 40 evaluation episodes.}
\label{tab:ablation}
\small
\setlength{\tabcolsep}{3.5pt}
\begin{tabular}{@{}lcccccccc@{}}
\toprule
& \multicolumn{3}{c}{\textbf{Components}} & \multicolumn{2}{c}{\textbf{Performance}} & \multicolumn{3}{c}{\textbf{Cost}} \\
\cmidrule(lr){2-4}\cmidrule(lr){5-6}\cmidrule(l){7-9}
\textbf{Variant} & Rec. & FMs & Aug. & Q-score & Succ. & Tokens & GPU h & Human h \\
\midrule
RLC (no added data)                   & --     & --     & --     & 0.633 & 0.425 & -- & -- & -- \\
\midrule
Success-only data                     & \xmark & \xmark & \xmark & 0.642 & 0.350 & -- & 72.4 & 16 \\
Per-failure recovery                  & \cmark & \xmark & \xmark & 0.267 & 0.20 & 3,366M & 158 & 0 \\
Per-failure recovery $+$ aug.         & \cmark & \xmark & \cmark & 0.483 & 0.375 & 528M & 44 & 0 \\
FMs without augmentation              & \cmark & \cmark & \xmark & 0.758 & 0.50 & 206M & 147 & 0 \\
Full \ours{}                          & \cmark & \cmark & \cmark & 0.84 & 0.58 & 261M & 36 & 0 \\
\bottomrule
\end{tabular}

\end{table}

\subsection{Ablations}
\label{sec:ablation}

We ablate the components of \ours{} on the task \emph{picking up trash} in BEHAVIOR-1K (Table~\ref{tab:ablation}). This task is selected as a representative task that consists of long-horizon, repeated skills and in which multiple failure modes occur. As illustrated in Table~\ref{tab:ablation}, we compare variants trained from the same RLC checkpoint using different training data compositions. The variants differ in whether all failures of an FM share one detector and one recovery program (FMs) and in whether failure states are multiplied by failure-conditioned augmentation (Aug.). Success-only data, which consists of successful demonstrations that do not target failures, serves as an untargeted control. All variants generate recovery data with the same agent and with approximately the same total length of 9 to 10~h, matching the total length of the original training data at a 1:1 ratio. Each fine-tuned policy is then evaluated on 40 test episodes. Appendix~\ref{app:ablation-details} describes the implementation of the variants and the cost accounting.

\paragraph{Targeted versus untargeted data.}
Full \ours{} increases the Q-score of the base policy from 0.633 to 0.84, whereas Success-only data only reach a Q-score of 0.642 while requiring 72.4 GPU hours and approximately 16 hours of human work.
Recovery data targeted at the failures of the policy are therefore more effective than a comparable amount of untargeted demonstrations, and they require half the GPU hours and no human effort. The importance of data thus comes not only from diversity but, most importantly, from distributions focusing on the weaknesses where the policies are likely to fail.

\paragraph{Shared failure-mode programs.}
Per-failure recovery, in which a separate agent localizes and recovers each failed rollout with its own program, reduces the Q-score of the base policy to 0.267. We attribute this degradation to the inconsistency of the resulting demonstrations, which motivates the shared FM programs of \ours{} (Section~\ref{sec:evolution}). BEHAVIOR-1K tasks require high-dimensional mobile manipulation, in which a 6-D end-effector pose can be reached by many configurations of the 7-DoF arm, the 4-DoF trunk and the mobile base. Independent agents relying on motion planners therefore admit many different solutions for recovering similar failure modes, and the resulting data exhibit few recurring patterns. Such high-entropy demonstrations have been shown to be unfavorable for policy learning~\citep{zhu2025sampling}. In FMs without augmentation, with one detector and one consistent recovery program per FM, the Q-score rises from 0.267 to 0.758, approaching the 0.84 of full \ours{}. Sharing the programs also reduces agent token usage by a factor of 16. The cost of per-failure recovery grows with the number of failures because each failure requires its own agent session, whereas the FM programs are written once from a small sample of rollouts and then executed on all detected failure states without the agent.

\paragraph{Failure-conditioned augmentation.}
The main benefit of augmentation is a lower simulation cost. In per-failure recovery with augmentation, the agent adapts the program written for each failure to several perturbed copies of its failure state, which requires fewer tokens than solving a new failure from scratch. The same amount of data then requires far fewer failures and thus fewer rollouts, which reduces agent token usage by a factor of six and simulation by a factor of 3.6. The saved time comes from the fact that, without augmentation, the FM detectors must process a much larger set of rollouts to collect enough failure states, whereas full \ours{} obtains its failure states from 50 rollouts and perturbs them, reducing simulation time from 147 to 36 GPU hours. Augmentation also improves the resulting policies, raising the Q-score from 0.267 to 0.483 for per-failure recovery and from 0.758 to 0.84 with FM programs. Since the perturbed states of a failure are recovered by the same program or an adaptation of it, their recoveries remain similar and the data become more consistent. These results indicate that, on a high-dimensional mobile manipulator, such consistent data facilitate the learning of complex skills.

\endgroup

\section{Conclusion and Future Work}

In this paper, we presented \ours{}, a failure-guided co-evolution framework that couples the refinement of an agent's executable knowledge with the post-training of robotic foundation models. In each round, the agent identifies the recurring failure modes of the current policy and develops or refines a reusable detector and recovery program for each mode. By combining these detectors and recovery programs with failure-conditioned augmentation, \ours{} generates corrective experience targeted at the current policy's weaknesses. Post-training on this experience expands the capabilities of the policy, and the failures of the updated policy in turn guide the next round of agent refinement. Experiments on BEHAVIOR-1K, RoboTwin, and RoboCasa show that \ours{} improves task performance over successive co-evolution rounds starting from strong initial policies and reduces the frequency of failures caused by targeted failure modes. Ablation studies further evaluate the contributions of individual components to policy performance and data acquisition efficiency.

\subparagraph{Limitation.}Currently, \ours{} learns only from successful recovery demonstrations, while the VLA's own successful and failed rollouts are not used for training. Future work could combine \ours{} with reinforcement learning to leverage these rollouts, improving data utilization through outcome-based credit assignment.

\bibliographystyle{plainnat}
\bibliography{references}

\clearpage
\appendix

\section{RoboCasa Tasks}
\label{app:robocasa-tasks}

Table~\ref{tab:robocasa-tasks} lists the ten RoboCasa tasks used in our experiments: four atomic tasks and six composite tasks from different activities.

\begin{table}[h]
\centering
\small
\setlength{\tabcolsep}{4pt}
\renewcommand{\arraystretch}{1.15}
\caption{\textbf{RoboCasa tasks used in our experiments.} Instructions are the default language instructions of each task; terms in angle brackets are sampled per episode.}
\label{tab:robocasa-tasks}
\begin{tabular}{@{}l >{\raggedright\arraybackslash}p{0.2\linewidth} >{\raggedright\arraybackslash}p{0.5\linewidth}@{}}
\toprule
\textbf{Task} & \textbf{Type (activity)} & \textbf{Instruction} \\
\midrule
TurnOnSinkFaucet & Atomic & Turn on the sink faucet. \\
TurnOffStove & Atomic & Turn off the $\langle$knob$\rangle$ burner of the stove. \\
CoffeeSetupMug & Atomic & Pick the mug from the counter and place it under the coffee machine dispenser. \\
NavigateKitchen & Atomic & Navigate to the $\langle$fixture$\rangle$. \\
\midrule
KettleBoiling & Composite (brewing) & Pick the kettle from the counter and place it on a stove burner. Then turn the burner on. \\
ArrangeTea & Composite (brewing) & Pick the kettle from the counter and place it on the tray. Then pick the mug from the cabinet and place it on the tray. Then close the cabinet doors. \\
LoadDishwasher & Composite (washing dishes) & Pick up the $\langle$dish$\rangle$ and $\langle$dish$\rangle$ from the counter, place them in the dishwasher, and close the dishwasher door. \\
SteamInMicrowave & Composite (steaming food) & Pick the $\langle$vegetable$\rangle$ from the sink and place it in the bowl. Then pick the bowl and place it in the microwave. Then close the microwave door and press the start button. \\
WashFruitColander & Composite (washing produce) & Put the colander in the sink, put the $\langle$fruits$\rangle$ in the colander, and turn on the sink faucet and pour water over the colander. \\
MakeIceLemonade & Composite (adding ice to beverages) & Grab a lemon wedge from the fridge and one ice cube from the ice bowl, and put them in the glass of lemonade. \\
\bottomrule
\end{tabular}
\end{table}

\section{Training Parameters}
\label{app:training}

\paragraph{BEHAVIOR-1K.} We initialize each task from the released per-task checkpoint of Robot
Learning Collective, the first-place solution of the 2025 BEHAVIOR Challenge, and each
round fine-tunes all parameters of the previous round's policy.
Each round trains for two epochs or 30,000 steps, whichever is shorter, with a global
batch size of 128, sampling the task's
200 human demonstrations and the accumulated recovery data at a $1{:}1$ ratio.
We use AdamW with $\beta = (0.9, 0.95)$, a weight decay of $10^{-10}$ and gradient
clipping at 1.0, with an EMA decay of 0.99. The learning rate is warmed up linearly
over 1,000 steps to a peak of $5 \times 10^{-5}$, followed by cosine decay to $10^{-6}$.
Action targets are normalized with the base checkpoint's statistics and clipped to $\pm 10$.

\paragraph{RoboTwin.}
We initialize $\pi_{0.5}$ from the checkpoint trained on both clean and randomized RoboTwin demonstrations
for round 1 and from round 1's final EMA checkpoint for round 2,
resetting the optimizer.
Each round fine-tunes all parameters for 40,000 optimizer updates
with a global batch size of 128, sampling original demonstrations
and accumulated recovery data at a $1{:}1$ ratio.
We use AdamW with $\beta=(0.9,0.95)$, a weight decay of $10^{-10}$,
and an EMA decay of 0.9.
The learning rate is warmed up linearly over 500 steps to a peak
of $2\times10^{-5}$, followed by cosine decay to $2\times10^{-6}$.

\paragraph{RoboCasa.} 
We initialize XR-1 from the released Xiaomi-Robotics-1-RoboCasa365 checkpoint.
Each fine-tuning stage consists of 30,000 optimizer updates with a global batch size of 192.
We use AdamW with a weight decay coefficient of $0.1$ and a peak learning rate of
$2 \times 10^{-5}$, with 500 linear warmup steps followed by cosine decay
to $5 \times 10^{-6}$.
Within each run, we select the checkpoint with the highest success rate
on a 200-episode development set and evaluate it on a 500-episode test set.

\section{Ablation Details}
\label{app:ablation-details}
\paragraph{Data budget and variants.}
We use picking up trash (BEHAVIOR-1K task~1), in which the robot must place three soda cans into a trash can. All variants start from the RLC checkpoint and draw their failures from its rollouts on this task, excluding the 20 public test instances. Each variant uses only as many of these rollouts as it needs to generate its data, and its GPU cost is computed from the rollouts it actually uses. Each variant generates an amount of new data comparable to the original demonstrations (9.76~h), and these data are sampled 1:1 with the original demonstrations during fine-tuning. Apart from the ablated component, the pipeline is identical to that of the main experiments. Failure states are restored from the simulator states recorded in the rollouts, and recovery programs act through the action space of the robot using classical methods, namely inverse kinematics, interpolation, navigation and closed-loop correction. The variants are as follows (FMs: shared failure-mode programs, Aug.: failure-conditioned augmentation).
\begin{itemize}
\item \textbf{Success-only data.} Approximately 5~h of successful rollouts of the base policy and 5~h of demonstrations collected by human teleoperation, which cost 72.4 GPU hours and approximately 16 hours of human work, respectively. These data follow the original data distribution and do not target failures.
\item \textbf{Per-failure recovery.} Each failed rollout is handled by a separate agent, which identifies the failed subtask, restores start states shortly before or after the error, and writes a program that completes this subtask. A recovery succeeds when the goal condition of the subtask is satisfied. Neither FMs nor augmentation are used.
\item \textbf{Per-failure recovery with augmentation (Aug.).} Per-failure recovery on 50 randomly chosen failures, whose failure states are multiplied by failure-conditioned augmentation. The agent adapts the program of each failure to its perturbed states, which requires fewer tokens than solving a new failure from scratch.
\item \textbf{FMs without augmentation (FMs).} The five FMs of full \ours{} on this task and their recovery programs, with detectors implemented from the same FM specifications. The detectors locate the preventive and corrective states of each FM in the failed rollouts, and the recovery program of the FM is executed on these states without augmentation.
\item \textbf{Full \ours{} (FMs $+$ Aug.).} The first round of the main experiments on this task (Section~\ref{sec:method}): five FMs discovered from 50 rollouts, their detectors and recovery programs, and failure-conditioned augmentation of the detected states.
\end{itemize}

\paragraph{Cost accounting.}
\emph{Tokens} are the sum of input, cache-write, cache-read and output tokens over all agent sessions of a variant and over the orchestrating session that launched and monitored them. A variant that reuses programs of full \ours{} also includes the tokens of the agents that wrote them. \emph{GPU hours} include only the simulation that produces data,
{\begin{equation*}
G = G_{\mathrm{roll}} + G_{\mathrm{rec}}, \qquad G_{\mathrm{roll}} = N_{\mathrm{roll}}\,\bar{t},
\end{equation*}}%
where $G_{\mathrm{roll}}$ is the cost of the base-policy rollouts that a variant uses to find its failure states, $N_{\mathrm{roll}}$ is the number of these rollouts, $\bar{t} = 0.25$ is the average number of GPU hours per rollout, and $G_{\mathrm{rec}}$ is the cost of recording the recovery data (Table~\ref{tab:ablation-gpu}). Runs for development, debugging and testing are excluded for all variants. Since recording requires approximately the same time for the same amount of data, the differences in GPU hours between variants arise mainly from the number of rollouts each variant needs to obtain enough failure states. All costs are reported for the same amount of data, namely the 9.06~h of recovery data of full \ours{}. Costs that increase with the amount of data, such as the per-failure agent sessions, the rollouts and the recording, are rescaled to this amount, whereas one-time costs, such as the orchestrating sessions and the programs reused from full \ours{}, are counted once. Recording time is additionally converted to the simulator throughput of full \ours{}, so that all variants are compared on the same hardware.

\begin{table}[h]
\centering
\caption{\textbf{Composition of the GPU hours in Table~\ref{tab:ablation}}, for 9.06\,h of recovery data. $N_{\mathrm{roll}}$: number of base-policy rollouts used to find the failure states. $G_{\mathrm{roll}} = N_{\mathrm{roll}} \bar{t}$, with an average of $\bar{t} = 0.25$ GPU hours per rollout. $G_{\mathrm{rec}}$: GPU hours for recording the recovery data at the simulator throughput of full \ours{}.}
\label{tab:ablation-gpu}
\small
\begin{tabular}{@{}lrrrr@{}}
\toprule
Variant & $N_{\mathrm{roll}}$ & $G_{\mathrm{roll}}$ & $G_{\mathrm{rec}}$ & Total \\
\midrule
Per-failure recovery & 545 & 136.3 & 21.5 & 158 \\
Per-failure recovery $+$ aug. & 87 & 21.8 & 22.3 & 44 \\
FMs without augmentation & 499 & 124.8 & 22.3 & 147 \\
Full \ours{} & 50 & 12.5 & 23.3 & 36 \\
\bottomrule
\end{tabular}
\end{table}

\section{Failure-Mode Metrics}
\label{app:fm-metrics}

The failure-mode columns of Table~\ref{tab:main} are computed over the test episodes $E$ of a policy on a benchmark, of which $F \subseteq E$ fail. Occurrences and recoveries are determined by the agent-written detectors on the recorded scene states: an occurrence counts as recovered if the policy, without assistance, later completes the subtask in which it occurred, i.e., the subtask's completion condition (Section~\ref{sec:discovery}) holds before the episode ends. A failure mode is \emph{unrecovered} in an episode if it occurs and the policy never recovers from it.
The base and first-round policies are evaluated using first-round detectors, with targeted FMs defined as those that received recovery data in that round. The second-round policy is evaluated using revised and new detectors, alongside first-round detectors for unchanged FMs. We report the following failure analysis metrics:
\begin{itemize}
    \item \emph{FM occurs} $= |E_{\mathrm{occ}}| / |E|$, where $E_{\mathrm{occ}}$ is the set of episodes in which at least one targeted FM occurs.
    \item \emph{Self-recovered} $= R / N$, where $N$ is the number of occurrences of targeted FMs over all episodes and $R$ the number of those the policy recovers from on its own.
    \item \emph{Fail $\mid$ FM} $= |F_{\mathrm{tgt}}| / |E_{\mathrm{occ}}|$, where $F_{\mathrm{tgt}} \subseteq F$ is the set of failed episodes with an unrecovered targeted FM.
    \item \emph{Targeted FM} $= |F_{\mathrm{tgt}}| / |E|$.
    \item \emph{Unexplained} $= |F_{\mathrm{none}}| / |E|$, where $F_{\mathrm{none}} = F \setminus F_{\mathrm{tgt}}$ is the set of failed episodes with no unrecovered targeted FM.
\end{itemize}

\section{Per-Failure-Mode Analysis}
\label{app:fm-level}

\begin{figure}[!htbp]
\centering
\includegraphics[width=0.85\linewidth]{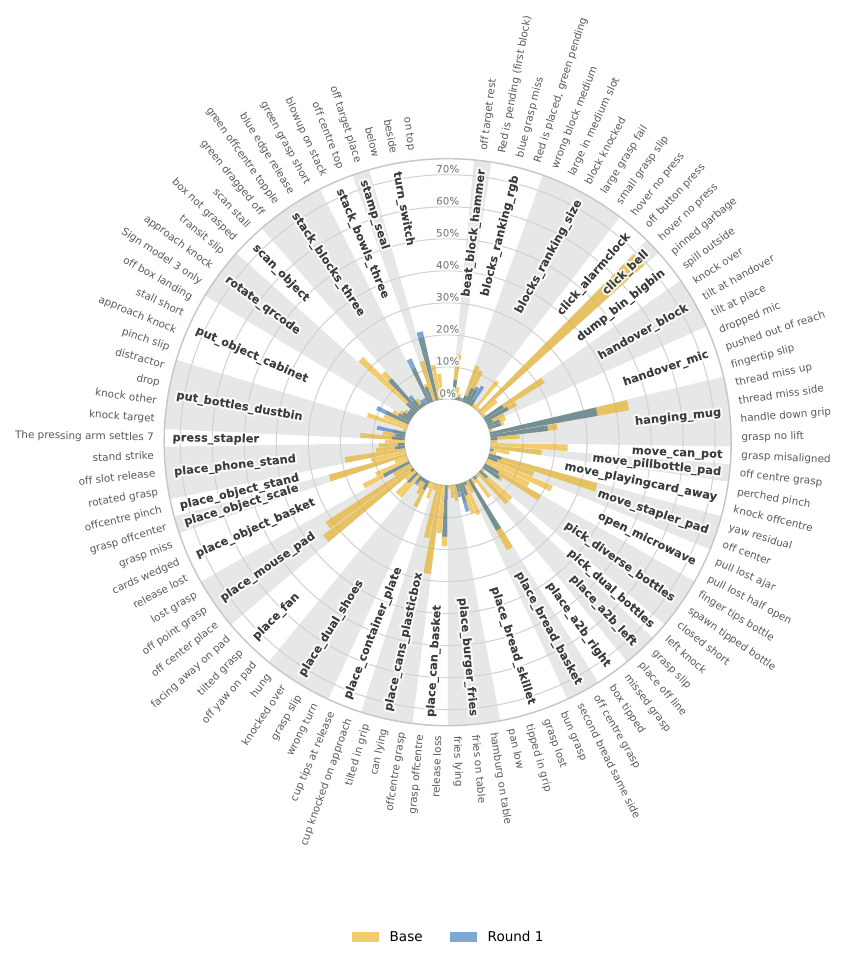}
\caption{\textbf{Per-FM failure rates on RoboTwin, first round.} Base vs.~first-round policy, first-round detectors. Each spoke is a targeted FM, and shaded sectors group the FMs by task. Bars give the fraction of the task's test episodes that fail with the FM never recovered from, before (yellow) and after (blue) the round; gray marks their overlap.}
\label{fig:fm-radar-robotwin}
\end{figure}

\begin{figure}[p]
\centering
\includegraphics[width=0.85\linewidth]{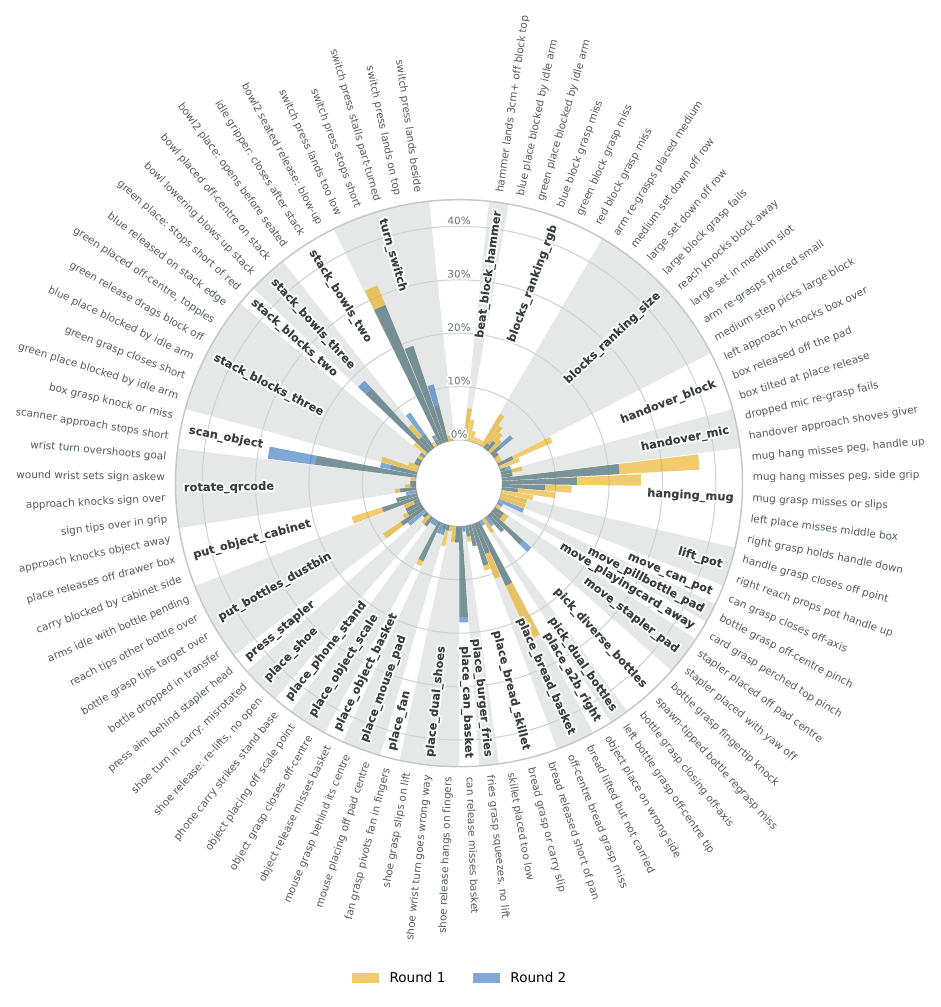}
\caption{\textbf{Per-FM failure rates on RoboTwin, second round.} First-round vs.~second-round policy, detectors after the second round of co-evolution. Same format as Figure~\ref{fig:fm-radar-robotwin}.}
\label{fig:fm-radar-robotwin-r2}
\end{figure}

\begin{figure}[p]
\centering
\includegraphics[height=0.44\textheight,keepaspectratio]{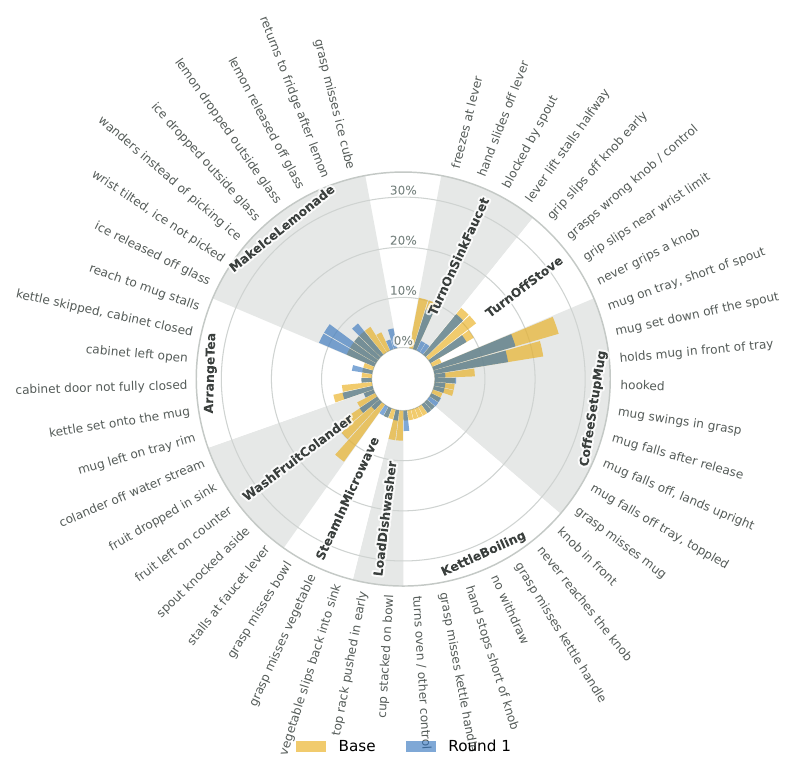}\\[2pt]
{\small (a) Round~1: base vs.\ first-round policy.}\\[10pt]
\includegraphics[height=0.44\textheight,keepaspectratio]{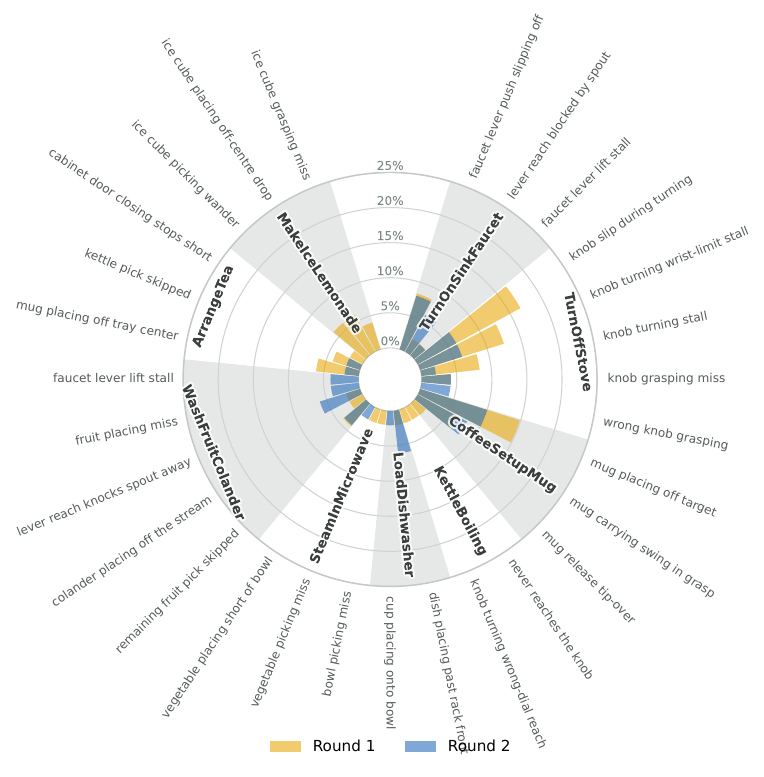}\\[2pt]
{\small (b) Round~2: first-round vs.\ second-round policy.}
\caption{\textbf{Per-FM failure rates on RoboCasa (XR-1).} Same format as Figure~\ref{fig:fm-radar-robotwin}. Round~1 uses the first-round detectors and Round~2 the detectors after the second round.}
\label{fig:fm-radar-robocasa}
\end{figure}

\begin{figure}[p]
\centering
\includegraphics[height=0.44\textheight,keepaspectratio]{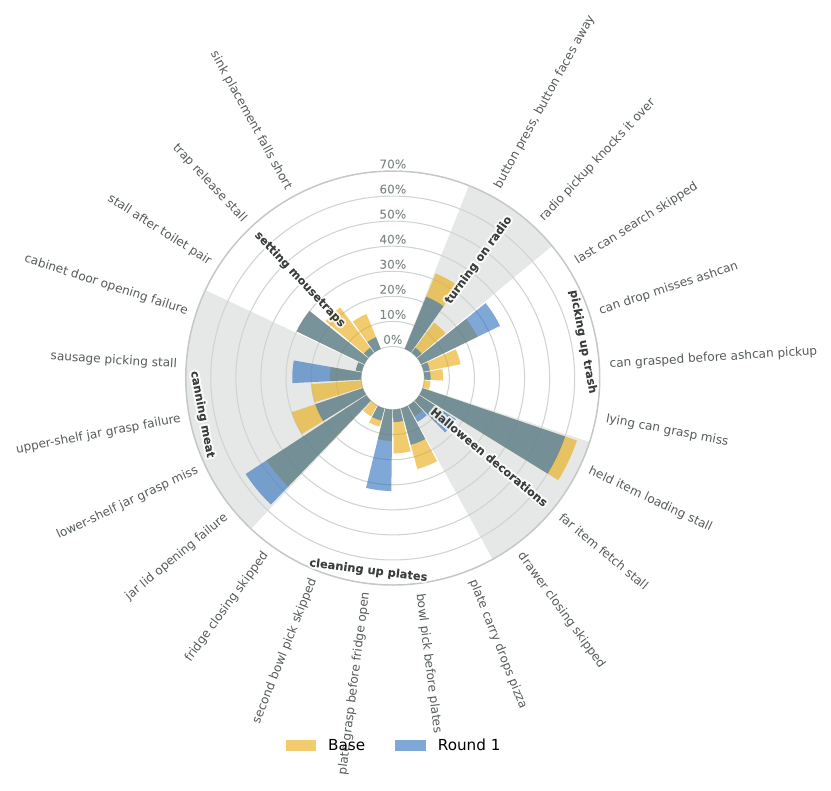}\\[2pt]
{\small (a) Round~1: base vs.\ first-round policy.}\\[10pt]
\includegraphics[height=0.44\textheight,keepaspectratio]{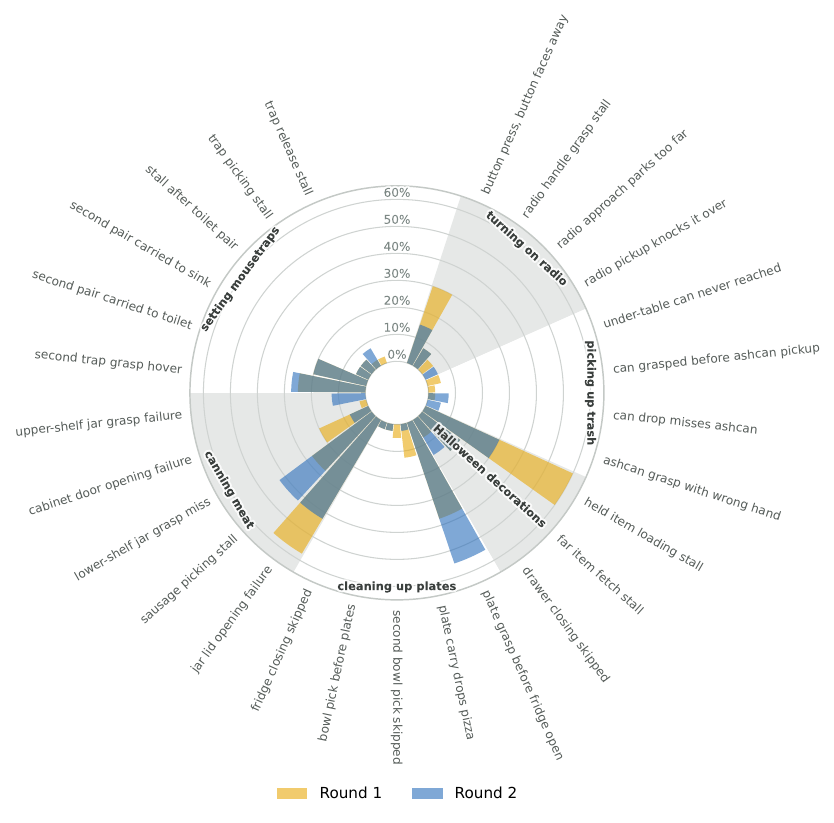}\\[2pt]
{\small (b) Round~2: first-round vs.\ second-round policy.}
\caption{\textbf{Per-FM failure rates on BEHAVIOR-1K, tasks 0 to 5.} Same format as Figure~\ref{fig:fm-radar-robotwin}. Round~1 uses the first-round detectors and Round~2 the detectors after the second round.}
\label{fig:fm-radar-b1k}
\end{figure}

Table~\ref{tab:main} aggregates the targeted failure modes (FMs) of each benchmark. Figures~\ref{fig:fm-radar-robotwin}--\ref{fig:fm-radar-b1k} break this aggregate down by FM, using the same detectors and counting rule. We compare the policies before and after each round of co-evolution. For example, after the first round of co-evolution, we compare the base policy with the first-round policy under the detectors written in the first round. For each FM, we report the fraction of its task's test episodes that fail with this FM never recovered from. This rate falls when the policy avoids the FM and also when it recovers after the FM occurs. FMs that cause no failure under either policy are omitted.

In the first round, failure rates decrease for 84 of the 98 FMs on RoboTwin, 29 of the 48 on RoboCasa, and 14 of the 22 on BEHAVIOR-1K. Among the FMs that cause at least 5\% of their task's episodes to fail under the base policy, the corresponding counts are 61 of 66, 16 of 20, and 13 of 20, respectively. The largest reductions occur for the FMs that dominate the base policy's failures: ``hovering over the bell without pressing it'' in \emph{click bell} on RoboTwin (69\% to 0\%), ``stall at the faucet lever'' in \emph{WashFruitColander} on RoboCasa (14\% to 0\%), and ``holding the traps without releasing them'' in \emph{setting mousetraps} on BEHAVIOR-1K (22.5\% to 2.5\%).

In the second round, failure rates decrease for 50 of the 86 FMs on RoboTwin and 19 of the 29 on RoboCasa. Among FMs meeting the same 5\% threshold under the first-round policy, the corresponding counts are 20 of 27 and 8 of 9, respectively. The largest reductions occur for ``hanging the mug with its handle up'' in \emph{hanging mug} on RoboTwin (37\% to 22\%), ``knob slipping during turning'' in \emph{TurnOffStove} on RoboCasa (17\% to 6\%), and both ``wandering instead of picking the ice'' and ``dropping the ice cube off the glass'' in \emph{MakeIceLemonade} (each 6\% to 0\%). On BEHAVIOR-1K, the largest second-round reductions occur for ``stalling with full hands at the shut drawers'' in \emph{putting away Halloween decorations} (60\% to 30\%) and ``pressing the button while it faces away'' in \emph{turning on a radio} (30\% to 15\%).

In Figure~\ref{fig:fm-radar-b1k}, the FMs of \emph{canning meat} remain frequent after both rounds, but the FMs of the earlier subtasks decline in each round, namely the jar-grasping failures in the first round and the lid-opening failure in the second, while the failures of the subsequent subtasks become more frequent. Canning meat shows how failures move to later subtasks in a long-horizon task. Its subtasks follow a fixed order: open the cabinet, take the jar out, open the jar's lid, fetch the sausage, and put it into the jar. Table~\ref{tab:canmeat-stages} groups the episodes by the furthest subtask they reach. Under the base policy, 51\% of episodes never get the jar out of the cabinet. After Round 1, this falls to 12\%, and opening the lid, which more episodes now reach, is exposed as the main bottleneck: 40\% of episodes get the jar out but never open the lid, up from 24\%. After Round 2, only 3\% of episodes stay at the jar and 18\% at the lid, while 79\% open the lid and reach the sausage, up from 24\% under the base policy; fetching the sausage becomes the next bottleneck. Each round thus moves the policy one subtask further. Because the Q-score credits only goal conditions satisfied at the end of an episode, it reflects this stage-wise progress only partially (0.02 to 0.09).

\begin{table}[H]
\centering
\small
\caption{\textbf{Progress through canning meat.} Fraction of episodes by the furthest subtask they reach.}
\label{tab:canmeat-stages}
\begin{tabular}{@{}lccc@{}}
\toprule
 & Base & Round 1 & Round 2 \\
\midrule
Jar not taken out of the cabinet & 0.51 & 0.12 & 0.03 \\
Jar taken out, lid not opened & 0.24 & 0.40 & 0.18 \\
Lid opened, sausage reached & 0.24 & 0.48 & 0.79 \\
\midrule
Q-score & 0.02 & 0.03 & 0.09 \\
\bottomrule
\end{tabular}
\end{table}

\end{document}

%% file: sections/intro.tex
\section{Introduction}
\label{sec:introduction}
\begin{figure}[t]
\centering
\includegraphics[width=0.98\linewidth]{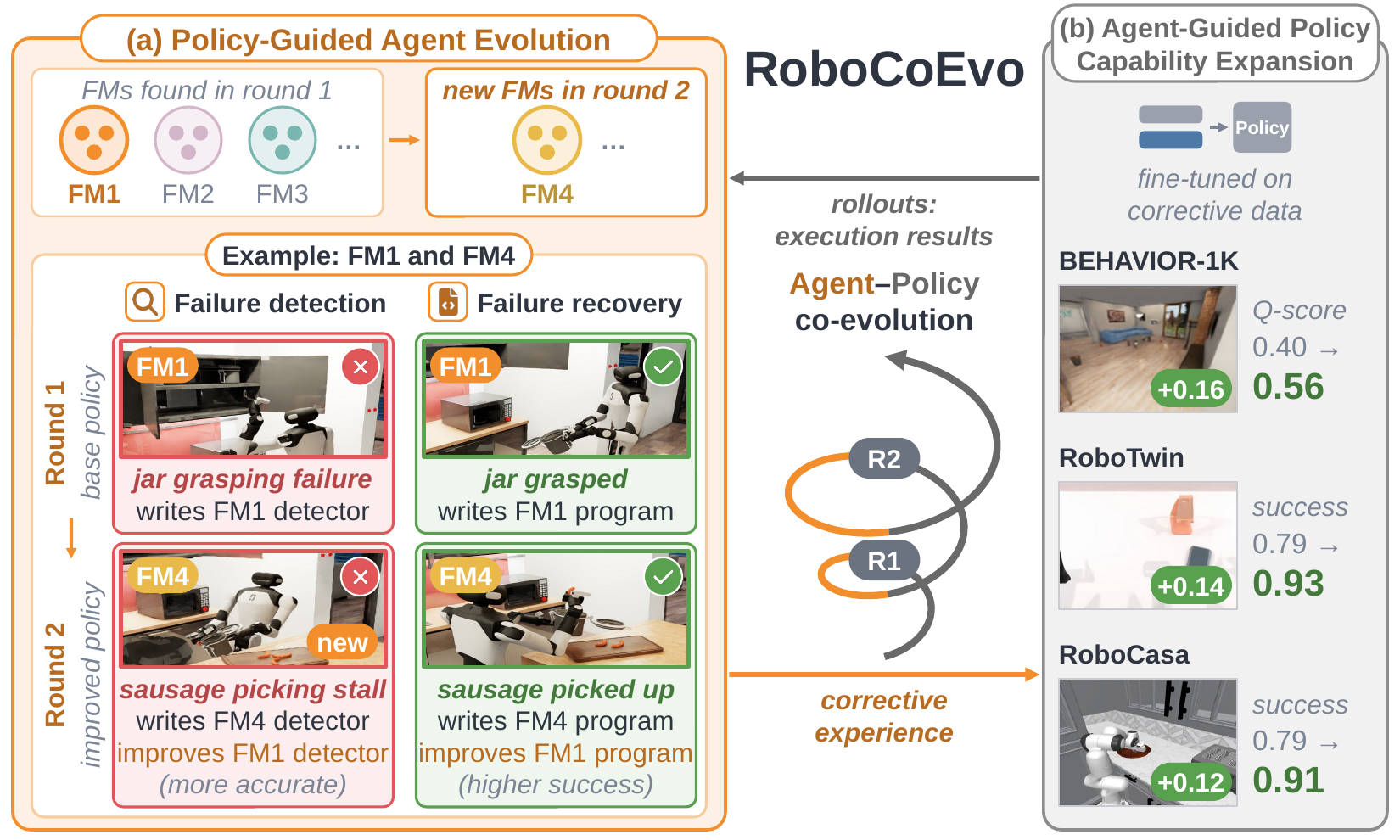}
\caption[]{\textbf{Overview of \ours{}.} The co-evolution framework of agent and policy improving each other over successive rounds.
(a) Policy-Guided Agent Evolution: the agent locates failures in the policy's failed rollouts and groups recurring failures
into failure modes (FMs), shown at the top for each round. For each FM, it writes a detector that identifies the failure
state in a rollout (red) and a recovery program that completes the failed subtask from this state in simulation (green).
As the policy improves, its rollouts expose FMs that did not occur in earlier rounds. In each subsequent round, the agent
writes detectors and recovery programs for the newly identified FMs and refines those of persisting FMs to improve
detection accuracy and recovery success. (b) Agent-Guided Policy Capability Expansion: the policy is fine-tuned on the
corrective experience together with the original demonstrations, which enables the policy itself to avoid or recover from the targeted
failures. The rollouts of the updated policy initiate the next round, and its task success increases over
successive rounds. The scores compare the base policy with the policy after \ours{}.}
\label{fig:teaser}
\end{figure}
Recent work has shown that MLLM-based agentic harnesses can enhance the capabilities of robotic foundation models~\citep{roboharness,zhang2026harness}. In these works, robotic foundation models provide visuomotor skills acquired from large-scale demonstrations~\citep{kim2024openvla,black2024pi0,black2025pi05}, while agents support their policy execution through failure diagnosis, recovery, and coordination~\citep{liu2023reflect,lin2025failsafe}. This complementary division of roles enables the combined system to address limitations of the robotic foundation models. 
However, the improvement processes of the agent and the robotic foundation model in the combined system typically remain separate or only loosely coupled~\citep{zhang2026robocop,yu2026coda}, with advances in one component not necessarily benefiting the other.
Without learning from corrective experience, the policy may remain dependent on agent assistance in similar situations. Meanwhile, an agent whose knowledge is not refined as the policy is updated may continue to target weaknesses that no longer constrain performance. Sustained improvement therefore requires a co-evolution mechanism that connects policy learning with the adaptation of agent assistance.

For corrective interventions to yield lasting policy improvements, the policy must incorporate the demonstrated behaviors into its own capabilities. 
Agent interventions provide training experience at states where execution fails, including states induced by accumulated errors and underrepresented in the training distribution~\citep{ross2011dagger}. Interactive imitation learning obtains such corrections from human supervisors~\citep{kelly2019hgdagger,liu2022sirius}, while recent approaches automate recovery and generate corrective data for policy training~\citep{zhang2026lrs,liao2026vlamotor,lang2026embodirsi}. Training on this experience allows the policy to internalize behaviors previously supplied by the agent. As these failures are resolved, however, the policy's need for assistance changes, and further improvement requires corrective experience tailored to its remaining limitations.

On the other hand, acquiring this experience requires the agent to adapt alongside the policy. It must reassess the updated policy's failures, determine where intervention is still needed, and revise its recovery strategies accordingly. Automated failure analysis can support this reassessment~\citep{duan2024aha,yu2026retrace}, while a coding agent can use execution feedback to refine the corresponding detectors and recovery programs. These revisions improve the agent's assistance and provide relevant corrections for the next policy update. However, applying conventional code-as-policy approaches without accounting for task-specific failure characteristics may compromise the accuracy of failure diagnosis and the efficiency of recovery, thereby limiting subsequent policy improvement. The two directions of improvement therefore form a continuous feedback loop: the robotic foundation model learns from the agent's interventions, and its remaining failures guide what the agent must diagnose and resolve next.

We present \ours{} (Figure~\ref{fig:teaser}), a co-evolution framework that couples \textbf{Policy-Guided Agent Evolution} with \textbf{Agent-Guided Policy Capability Expansion}. The agent evolves through revisions to executable failure-detection and recovery programs.  
Guided by policy rollouts and execution feedback, it identifies recurring failure modes, develops and evaluates recovery strategies, and generates corrective demonstrations across varied task configurations using a code-as-policy formulation~\citep{liang2022code}. The robotic foundation model then incorporates these corrective behaviors into its end-to-end policy through post-training. Subsequent evaluation of the updated model identifies residual capability gaps and directs the next round of program refinement. This iterative exchange improves the agent's diagnosis and recovery strategies while expanding the policy's capacity for independent execution. We instantiate and evaluate \ours{} with vision-language-action (VLA) models on BEHAVIOR-1K, RoboCasa, and RoboTwin. Experiments demonstrate sustained improvement across successive cycles, with higher autonomous task success and stronger generalization to unseen configurations. In summary, our contributions are:

\begin{enumerate}
    \item \textbf{Failure-Mode-Centric Agent--Policy Co-Evolution:} We introduce RoboCoEvo, a failure-guided co-evolution framework that couples the refinement of executable agent knowledge with learning-based policy post-training. Through a recursive feedback loop, the general agent generates corrective experience to improve the policy, while the evolving policy reveals new capability gaps that guide further agent adaptation.

    \item \textbf{Policy-Guided Agent Evolution:} We develop a failure-mode-centric approach that organizes recurring policy failures into reusable executable detectors and recovery programs. As the policy evolves, the agent continuously refines programs for persistent failure modes and develops new ones for emerging failures, enabling its corrective skills to evolve alongside the policy.

    \item \textbf{Agent-Guided Policy Capability Expansion:} We introduce failure-conditioned augmentation to scale the corrective experience generated by reusable recovery programs. By learning from these targeted demonstrations, the policy progressively internalizes agent-generated recovery behaviors and expands its capabilities. 

    \item \textbf{Significant Policy Improvement and Data Efficiency:} Experiments across BEHAVIOR-1K, RoboTwin, and RoboCasa demonstrate consistent policy improvements over successive co-evolution iterations. Extensive analyses further highlight the effectiveness of reusable failure-mode programs and targeted data augmentation in improving task performance and data-acquisition efficiency.
\end{enumerate}

\par\begingroup
\dimen0=\dimexpr\pagegoal-\pagetotal\relax
\ifdim\dimen0<7\baselineskip\newpage\fi
\endgroup
\section{Related Work}
\label{sec:related}

\paragraph{Agent-assisted robotic execution.}
LLM- and MLLM-based agents have been widely employed to enhance robotic execution without modifying the parameters of the underlying policies. Such agents have been applied to the detection and explanation of execution failures~\citep{liu2023reflect,duan2024aha,yu2026retrace,zhou2025codeasmonitor} and to intervention for failure recovery~\citep{lin2025failsafe,wang2026roboduo}, while hierarchical frameworks rely on them to orchestrate low-level policies and skills~\citep{shi2025hirobot,team2025geminirobotics15,roboharness,zhang2026harness}. As execution failures are difficult to anticipate, several systems further refine the agent's knowledge based on execution feedback, for instance by revising orchestration code~\citep{roboharness}, by accumulating executable skills~\citep{wang2023voyager,wen2026roborsi}, or by updating failure checks and recovery routines for failures grouped according to their underlying causes~\citep{ding2026zetta,sun2026reskill}. In these systems, however, the acquired improvements reside in the agent or its programs, whereas any learned policy responsible for low-level control retains fixed parameters. The capabilities of such a policy therefore remain unchanged, and the agent or its programs must compensate for its recurring failures at execution time.

\paragraph{Learning from corrective experience.}
Policies trained through imitation are prone to compounding errors that lead them to states underrepresented in their demonstrations~\citep{ross2011dagger}. Interactive imitation learning mitigates this distribution shift by training on corrections collected at the states visited by the policy~\citep{ross2011dagger,kelly2019hgdagger,liu2022sirius,hu2025rac}, and RECAP incorporates such corrections into multi-round training with a learned value function~\citep{amin2025recap}. Because these corrections are typically provided by human supervisors, whose effort scales with the number of failures, recent methods have sought to generate corrective data automatically. IntervenGen and FLARE extend a small number of human recovery demonstrations to new scene configurations and to new or perturbed failure states~\citep{hoque2024intervengen,zhao2026flare}, CARE uses a motion-planning oracle to annotate failure states sampled from a distribution fitted to failed rollouts~\citep{xiao2026care}, and other approaches derive corrections from fixed language-model prompts, pretrained experts, privileged continuations, or residual reinforcement-learning specialists~\citep{bernat2026magmagen,li2024scvla,snegirev2026kintsugi,xiao2025pld}. These studies demonstrate that training on failure-targeted corrections can improve the policy itself. Nevertheless, approaches that extend human demonstrations still rely on a supervisor to demonstrate each new recovery skill, while automatic generators are constructed once and are not refined from the failures they correct.

\paragraph{Coupled improvement of agents and policies.}
The works most closely related to ours couple an agent with policy training. LRS, VLAMotor, and EmbodiRSI use an MLLM agent or a motion planner to generate corrections from the failure states of a policy, on which the policy is subsequently fine-tuned~\citep{zhang2026lrs,liao2026vlamotor,lang2026embodirsi}. In these frameworks, however, each failure is corrected in isolation, so that the knowledge acquired in correcting one failure is neither retained nor refined for similar or subsequently emerging failures. Other frameworks improve the policy over multiple rounds but differ in how its corrective behaviors are acquired. Recova learns recovery behaviors with a separate recovery policy rather than with the task policy~\citep{liu2026recova}, and RoboClaw retrains its policies on their own rollouts and on human interventions under a fixed VLM controller~\citep{li2026roboclaw}. Robo-COP co-evolves the textual memory of an orchestrator with a policy but trains this policy exclusively on successful segments of its own executions, which restricts the learnable behaviors to those that its policy or scripted tools already partially exhibit~\citep{zhang2026robocop}. CoDA co-evolves a data engine with a policy by adapting its scene-generation and reward-shaping programs to the learning progress of this policy~\citep{yu2026coda}, and ENPIRE enables an agent to revise training recipes~\citep{xiao2026enpire}. In both cases, the corrective behavior is still obtained through the policy's own exploration or from demonstrations that do not originate at its failure states. SEAR outlines a comparable agent--policy loop as future work~\citep{wang2026sear}. These frameworks represent early efforts toward bidirectional improvement between agents and policies. Each, however, realizes only part of it. In some, the agent corrects failures one at a time without accumulating reusable knowledge of how to correct them, so that it adapts to individual failures but does not itself improve. In others, the agent and the policy are both updated, but the policy still learns mainly from its own executions or exploration, which can limit the recovery behaviors it acquires at the states where it fails. Effective co-evolution therefore requires that the agent accumulate and refine its knowledge against the policy's evolving failures, and that this knowledge be translated into targeted corrective demonstrations for the policy itself at the states where it fails.

%% file: sections/method.tex
\section{\texorpdfstring{\ours: Failure-Guided Agent--Policy Co-Evolution}{RoboCoEvo: Failure-Guided Agent-Policy Co-Evolution}}
\label{sec:method}

\subsection{Problem Setup and Overview}
\label{sec:problem}

\begin{figure}[t]
\centering
\includegraphics[width=\linewidth]{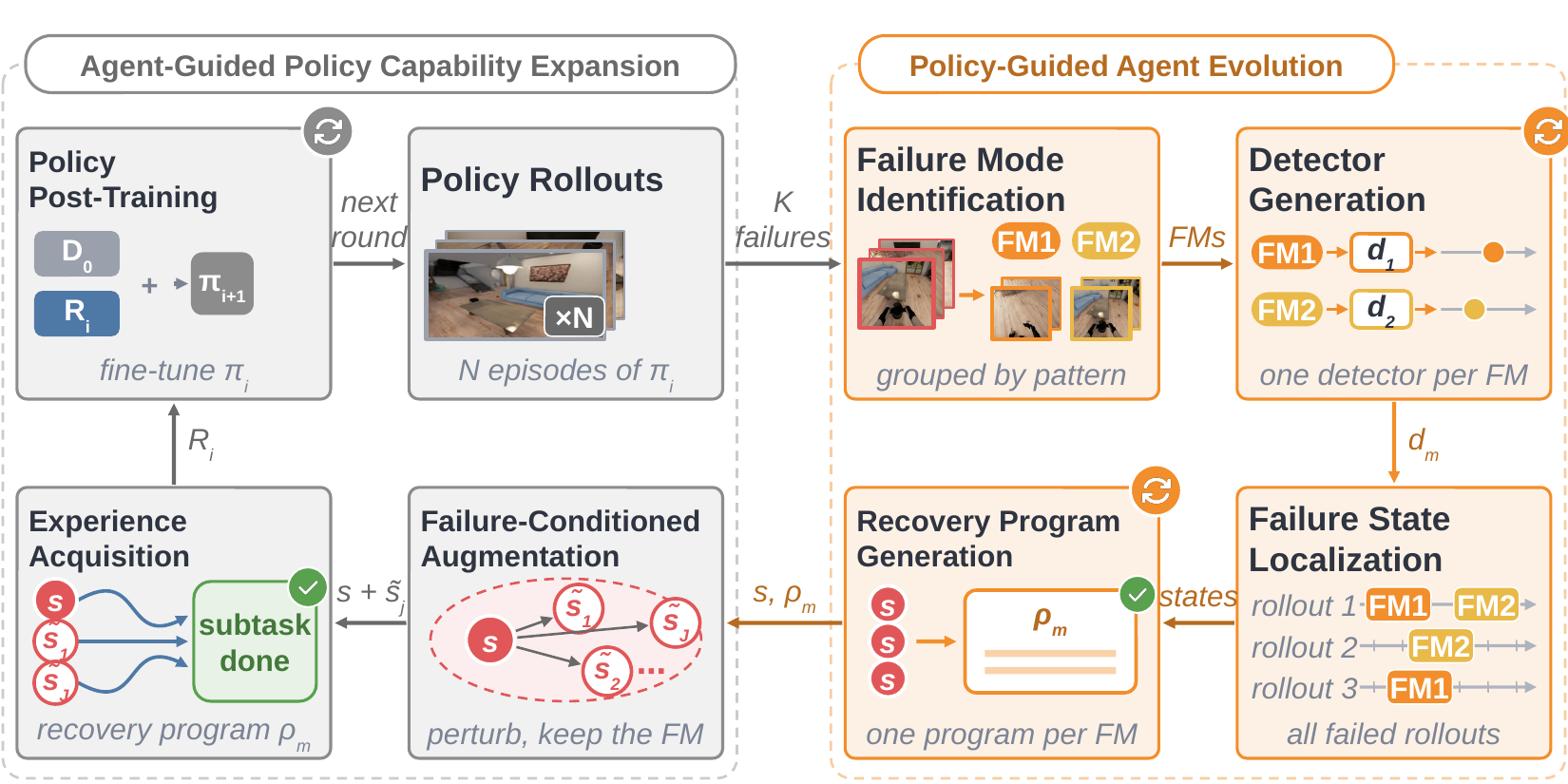}
\caption{\textbf{The \ours{} pipeline.} Each round proceeds clockwise, starting with rollouts of the current policy $\pi_i$. Policy-Guided Agent Evolution (right): the agent identifies failure modes (FMs) from $K$ sampled failures and writes a detector $d_m$ for each FM. The detectors localize failure states in all failed rollouts, from which the agent writes a recovery program $\rho_m$ for each FM. Agent-Guided Policy Capability Expansion (left): failure-conditioned augmentation perturbs each localized state $s$ into $J$ states $\tilde{s}_j$ of the same FM, and $\rho_m$ completes the failed subtask from both the original and the perturbed states, which yields the corrective experience $R_i$. The policy is then post-trained on $\mathcal{D}_0$ and the accumulated corrective experience to obtain $\pi_{i+1}$, whose rollouts start the next round. Steps marked with circular arrows are updated in every round.}
\label{fig:pipeline}
\end{figure}

As shown in Figure~\ref{fig:pipeline}, \ours{} formulates the co-evolution of an agent and a policy as an iterative process. The base policy $\pi_0$ is a robotic foundation model trained on reference demonstrations $\mathcal{D}_0$. In round $i$, \textbf{Policy-Guided Agent Evolution} (Section~\ref{sec:evolution}) collects rollouts of the current policy $\pi_i$ in a set of scenarios $\mathcal{S}$. An MLLM-based agent groups the recurring failures into \emph{failure modes} (FMs) and writes or revises a detector $d_m$ and a recovery program $\rho_m$ for each FM. \textbf{Agent-Guided Policy Capability Expansion} (Section~\ref{sec:update}) then applies these programs to the detected failure states and to perturbed copies of them, which yields the corrective experience
\begin{equation*}
    R_i = f(\pi_i, \mathcal{S}).
\end{equation*}
The policy is then post-trained on $\mathcal{D}_0$ and the corrective experience accumulated up to round $i$ to obtain $\pi_{i+1}$, whose rollouts are used in the next round.
\ours{} observes the scene primarily through a \emph{scene state} $s_t=(q_t,\{p^k_t\}_{k\in\mathcal{O}},\mathcal{F}_t,I_t)$, where $I_t$ denotes the camera images. Here, $q_t$ is the robot state, including the base pose and joint configuration, $\mathcal{O}$ is the set of task-relevant objects, and $p^k_t$ is the pose of object $k$. The \emph{symbolic state} $\mathcal{F}_t$ is the set of facts that hold at step $t$, such as $\mathrm{open}(k)$, $\mathrm{on}(k,k')$, and $\mathrm{held}(k)$. The agent derives and reconstructs these facts from the simulator state.

\subsection{\texorpdfstring{Policy-Guided Agent Evolution}{Policy-Guided Agent Evolution}}
\label{sec:evolution}
\label{sec:discovery}
\label{sec:acquisition}
In each round, the agent adapts its detectors and recovery programs to the failures of the current policy $\pi_i$. If the agent instead diagnosed and recovered each failed rollout separately, its cost would grow with the number of failures, and independent recoveries of similar failures could follow different strategies, leading to inconsistent demonstrations. \ours{} therefore groups recurring failures into FMs and writes one detector and one recovery program per FM. The cost of diagnosis and program writing is shared by all failures of an FM, and all recoveries of an FM follow the same strategy (Section~\ref{sec:ablation}).

\paragraph{Failure mode identification.}
To define a failure mode, we first define the \emph{failure pattern} $\mathcal{P}_m$ as a recurring pattern in sequences of scene states $(s_{t_f-h}, \ldots, s_{t_f+\ell})$ around the failure time $t_f$, where $h$ and $\ell$ denote the numbers of steps before and after the failure, respectively. Similar failure patterns are usually caused by similar root causes. For example, if a dual-arm robot is holding objects in both hands simultaneously, it cannot proceed to perform other manipulation tasks. Such cases can recur across different scenarios.

We further define \emph{subtasks} as successive stages of a task, each with a completion condition expressed in terms of symbolic facts. For example, a picking subtask is complete when $\mathrm{held}(k)$ holds, while a placement subtask is complete when $\mathrm{on}(k,k')$ holds. A long-horizon task can thus be divided into several subtasks, each of which can be treated as an independent capability to exercise or goal to achieve.

Hence, a \emph{failure mode} $m$ is defined as a group of failures that stall at the same subtask and exhibit the same \emph{failure pattern} $\mathcal{P}_m$. The failure pattern describes the root cause of the FM and the conditions under which the policy fails, primarily in terms of facts in $\mathcal{F}_t$ and their changes over time.

\paragraph{Detector generation and failure state localization.}
{The FM detector identifies occurrences of an FM in rollouts.} In each round, the agent reviews a sampled subset of failed rollouts to identify recurring failure modes and writes a detector for each mode. The generated detectors are then applied to all failed rollouts to locate matching instances. For each failure occurrence, the agent identifies a \emph{corrective state}, where the failure pattern first appears, and a \emph{preventive state}, an earlier state selected for demonstrating how to avoid the failure. The FM detector returns variable-size sets of preventive and corrective states:
\begin{equation*}
    d_m(\tau) = \big(
    \mathcal{S}_{\mathrm{pre}}^m(\tau),
    \mathcal{S}_{\mathrm{corr}}^m(\tau)\big),
\end{equation*}
where $\mathcal{S}_{\mathrm{pre}}^m(\tau)$ and
$\mathcal{S}_{\mathrm{corr}}^m(\tau)$ contain the preventive and corrective states identified for failure mode $m$ in rollout $\tau$, respectively. Each set may be empty or contain one or more states. The detector is implemented as a finite-state machine that identifies the failure pattern $\mathcal{P}_m$ offline from the recorded scene states of a rollout $\tau$. Within a round, the detectors are applied to all failed rollouts without further revision by the agent. In the next round, the agent revises them according to the failure modes of the updated policy.

\paragraph{Recovery program generation.}
For each FM, its detected preventive and corrective states are partitioned into development and held-out validation sets. The agent develops a shared failure recovery program $\rho_m$ using simulator states restored from the development set. Following a \emph{code-as-policy} approach, $\rho_m$ is implemented as a finite-state machine over $s_t$ that composes navigation and manipulation skills with planned motions~\citep{garrett2024skillmimicgen,sahoo2026datafarm}. Recovery success is defined as completion of the subtask at which the original execution failed. The program undergoes iterative refinement and is accepted only when its recovery success rate reaches at least $70\%$ on the held-out validation set. The recovery programs evolve alongside the policy. In each subsequent round, the agent develops new programs for newly identified FMs and refines existing programs for FMs that persist. These refinements aim to broaden the range of failure states that the programs can address, improve their recovery success rates, and more precisely target the weaknesses revealed by the current policy.

\subsection{\texorpdfstring{{Agent-Guided Policy Capability Expansion}}{Agent-Guided Policy Capability Expansion}}
\label{sec:update}
The detectors and recovery programs of Section~\ref{sec:evolution} are then used to generate corrective experience, on which the policy is post-trained.

\begin{figure}[t]
\centering
\includegraphics[width=\linewidth]{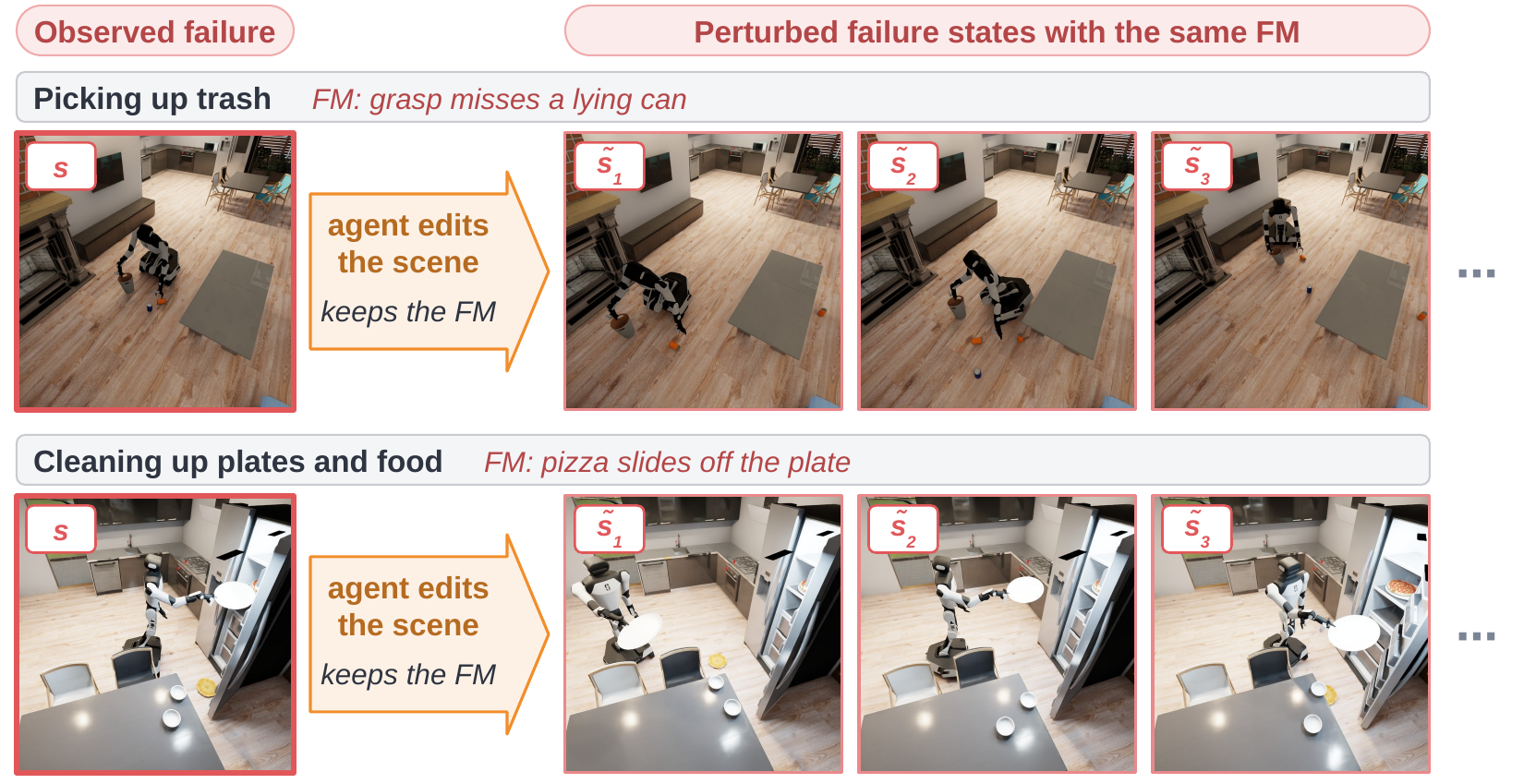}
\caption{\textbf{Failure-conditioned augmentation on two BEHAVIOR-1K tasks.} From an observed failure state
{$s$} (left), the agent edits the robot and object states in the simulator to obtain perturbed states
{$\tilde{s}_1,\tilde{s}_2,\ldots$} (right) that preserve the same FM. Each perturbed state serves as an additional initial
state for the recovery program, so a single observed failure yields multiple recovery runs.}
\label{fig:multiplication}
\end{figure}

\paragraph{Failure-conditioned augmentation and experience acquisition.}
\label{sec:multiplication}
To efficiently generate additional corrective experience covering the targeted failure mode, we introduce \emph{failure-conditioned augmentation}, which creates new failure states by perturbing observed ones in the simulator. This procedure is motivated by the cost of collecting additional failure instances through rollouts, as each rollout yields only a few instances of each FM.

For each FM, the agent identifies perturbable components of the scene state, including the robot state $q_t$, which encompasses the base pose, joint configuration, and end-effector poses, as well as the pose $p^k_t$ and symbolic facts in $\mathcal{F}_t$ associated with each task-relevant object $k$, regardless of whether the object is held. Additional failure scenes are then generated by perturbing these components while preserving the same failure mode. Figure~\ref{fig:multiplication} shows perturbed failure states on two BEHAVIOR-1K tasks. Perturbed preventive states must retain the conditions for demonstrating how to avoid the target failure, rather than already exhibit it.

The accepted recovery program $\rho_m$ is then executed from the detected failure states and from their perturbed copies without further agent intervention. Only successful recoveries are retained, and together they constitute the corrective experience $R_i$ of round $i$. Augmentation thus yields additional corrective experience from a limited number of observed failures, which reduces the number of rollouts required.

\paragraph{Policy post-training.}
The corrective experience $R_i$ is added to the training data together with the reference demonstrations, and the policy learns the corrective behaviors through DAgger-style post-training~\citep{ross2011dagger}:
\begin{equation}
\mathcal{D}_{i+1} = \mathcal{D}_0 \cup \bigcup_{j=0}^{i} R_j,
\qquad
\pi_{i+1} = \operatorname{FineTune}\big(\pi_i, \mathcal{D}_{i+1}\big).
\label{eq:train}
\end{equation}
During fine-tuning, we enforce a $1:1$ sampling ratio between the original reference demonstrations $\mathcal{D}_0$ and the accumulated corrective experience to mitigate catastrophic forgetting.
Because $\mathcal{D}_{i+1}$ contains the corrective experience of all previous rounds, behaviors learned in earlier rounds remain represented in the training data, and each new round adds experience for the remaining failures. The updated policy $\pi_{i+1}$ is deployed without the agent, and its rollouts are used in the next round of agent evolution.